\documentclass{article}
\usepackage{graphicx}		
\DeclareGraphicsExtensions{.pdf,.jpeg,.png}
\usepackage{color}
\usepackage{subfig}
\usepackage{amssymb, amsmath}                            
\usepackage{amsfonts}
\usepackage{mathtools}
\usepackage{pdfpages}
\usepackage{booktabs}
\usepackage{colortbl} 
\usepackage{xcolor} 
\usepackage{xfrac}
\usepackage{multirow}
\usepackage{tabularx}
\usepackage[linesnumbered,ruled]{algorithm2e}
\usepackage[numbers,square,sort&compress]{natbib}
\usepackage{authblk}
\usepackage{hyperref}
\hypersetup{
    colorlinks=true,
    linkcolor=blue,
    filecolor=magenta,      
    urlcolor=cyan,
    pdftitle={Overleaf Example},
    pdfpagemode=FullScreen,
    }

\usepackage{rotating}
\usepackage{tabularx}
\usepackage{multirow}
\usepackage{booktabs}

\usepackage{geometry}
\newgeometry{vmargin={1in}, hmargin={1in}}   
    
\usepackage[utf8]{inputenc}

\usepackage{adjustbox}

\makeatletter
\renewcommand\AB@affilsepx{, \protect\Affilfont}
\makeatother

\title{Surgical Visual Understanding (SurgVU) Dataset} 
\author{Aneeq Zia}
\author{Max Berniker$^*$}
\author{Rogerio Nespolo$^*$}
\author{Xiaorui Zhang$^*$}
\author{Conor Perreault$^*$}
\author{Ziheng Wang}
\author{Benjamin Mueller}
\author{Ryan Schmidt}
\author{Kiran Bhattacharyya}
\author{Xi Liu}
\author{Anthony Jarc}

\affil{Intuitive Surgical, Inc.}
\footnotetext{These authors contributed equally.}

\begin{document}

\maketitle

\begin{abstract}


Owing to recent advances in machine learning and the ability to harvest large amounts of data during robotic-assisted surgeries, surgical data science is ripe for foundational work.
We present a large dataset of surgical videos and their accompanying labels for this purpose. We describe how the data was collected and some of its unique attributes. Multiple example problems are outlined. Although the dataset was curated for a particular set of scientific challenges (in an accompanying paper), it is general enough to be used for a broad range machine learning questions. Our hope is that this dataset exposes the larger machine learning community to the challenging problems within surgical data science, and becomes a touchstone for future research. The videos are available at \url{https://storage.googleapis.com/isi-surgvu/surgvu24_videos_only.zip}, the labels at \url{https://storage.googleapis.com/isi-surgvu/surgvu24_labels_updated_v2.zip}, a validation set for tool detection problem at \url{https://storage.googleapis.com/isi-surgvu/cat1_test_set_public.zip}, and a sample set of question \& answer pairs dataset for surgical visual question answering at \url{https://storage.googleapis.com/isi-surgvu/SURGVU25_cat_2_sample_set_public.zip}.

\end{abstract}

\section*{Introduction}

Robotic-assisted surgery (RAS) is an increasingly popular choice, with well over a million cases performed each year \cite{Childers2018, mederos2022trends, Perez2019, grimsley2022exploring}, and larger future rates predicted \cite{Rizzo2023}. Aside from the many benefits over traditional surgery, RAS yields, as a matter of course, large amounts of data. High resolution endoscopic video data, precisely measured surgeon movements, and tool kinematics are some obvious examples. This data, exceedingly difficult to gather before RAS, is now par for the course and presents particularly auspicious opportunities for machine learning. 

The Surgical Visual Understanding dataset (abbreviated, SurgVU dataset) is a collection of surgical videos and attendant labels. The videos were captured during robotic surgery training sessions, hosted by Intuitive Surgical. In these training sessions surgeons perform a series of standard surgical steps on a porcine model. Information automatically harvested from the robotic device is used to provide presence labels for the installed tools. In total the dataset contains over 840 hours of video at 60 frames per second, for a total of approximately 18 million labeled images.

The video and labels are part of accompanying machine learning challenges \cite{eisenmann2022biomedical, eisenmann2023winner, zia2023surgical} hosted each year by Intuitive Surgical through the MICCAI conference as part of the Endoscopic Vision (EndoVis) challenge \cite{speidel2021endoscopic,zia2022endoscopic}. In this challenge, participating teams use the video dataset to work on cutting-edge problems in the field of surgical data science \cite{maier2020surgical} like surgical tool detection and step recognition. 
There are a host of research areas and applications outside this immediate domain that are also relevant (e.g. video segmentation, detection, and generative algorithms).

As with other public datasets \cite{allan2017, allan2018, twinanda2016endonet, ahmidi2017dataset, wagner2021comparative, zia2021surgical, zia2022objective}, this one can serve multiple purposes. First, it lowers the barrier for new, aspiring ML enthusiasts by offering a simple entry point for training models on large data. It can serve as a benchmark for algorithm development, allowing researchers to compare and contrast their work on a common point-of-reference dataset. And, it exposes researchers across diverse disciplines to the challenges of surgical data science. Finally, the overall effect of public datasets is to accelerate research, allowing for advances that far outpace what can be accomplished by a single research group.

To the best of our knowledge this is the largest publicly available surgical video dataset available. In the future we hope to expand it, perhaps by adding more video, or by including further labels. Importantly, as a public dataset there is the opportunity for the community to expand it in similar, and diverse ways we cannot anticipate. Ultimately we hope this dataset becomes a touchstone for the broader field of surgical data science.






\section*{Dataset}

\subsection*{Dataset overview \& composition}

\begin{figure}[ht!] 
    \centering
    \includegraphics[width=0.9\linewidth]{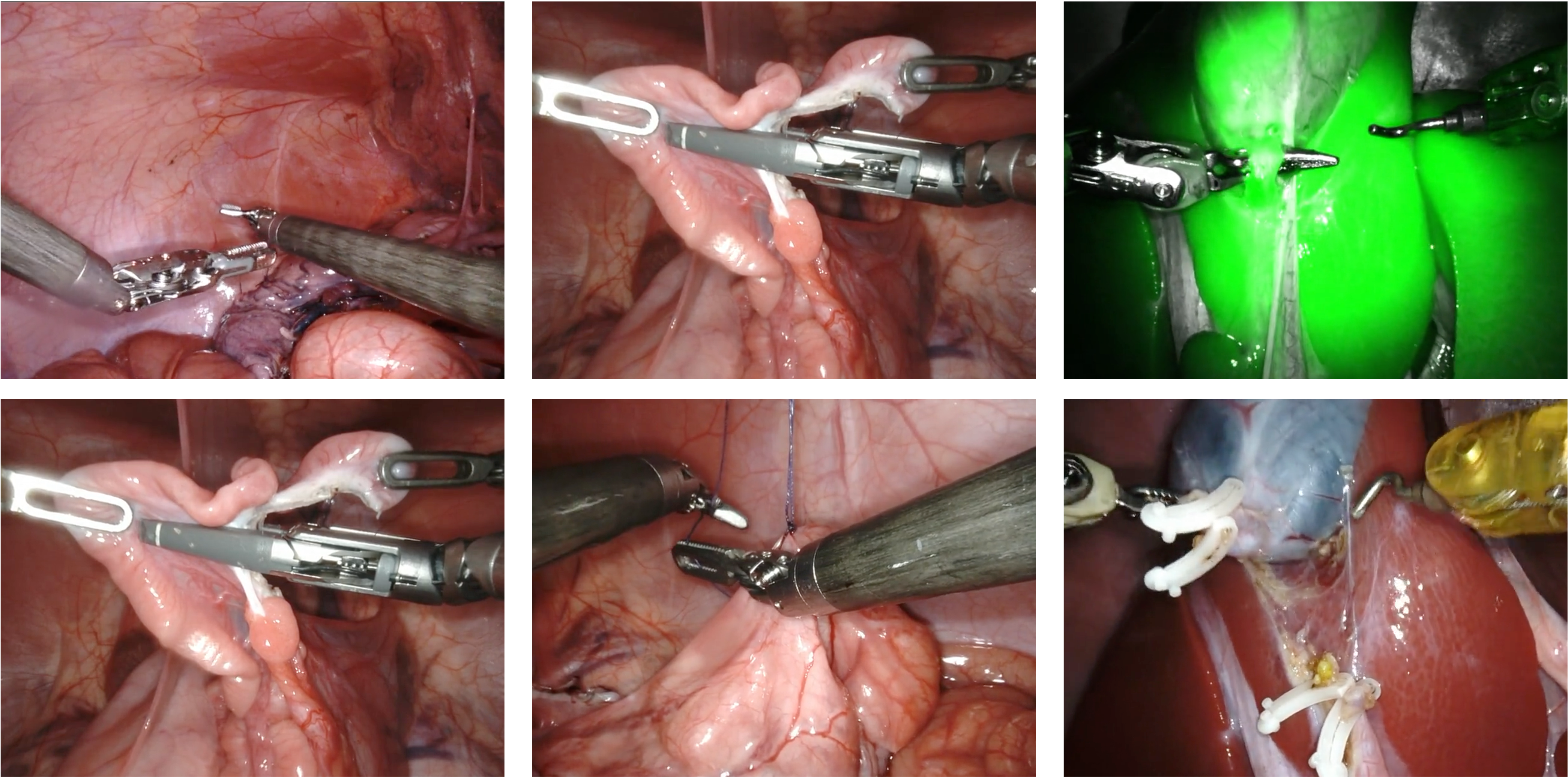}
    \caption{Sample frames of multiple surgical steps included in our dataset. Note that some frames capture fluorescence imaging (top right image).}
    \label{fig:example_video_frames}
\end{figure}

This dataset was built upon standardized surgical training steps in a controlled environment. The da Vinci robotic system was used to conduct training exercises, during which trainee and expert surgeons performed these steps on porcine tissues. The exercises were recorded using video and system data, with the captured video displaying the surgeon's view from the console (see Fig \ref{fig:example_video_frames}).\\

\noindent{\bf{Tools}}

\begin{figure}[tbh!] 
 \centering
 \includegraphics[scale=0.5]{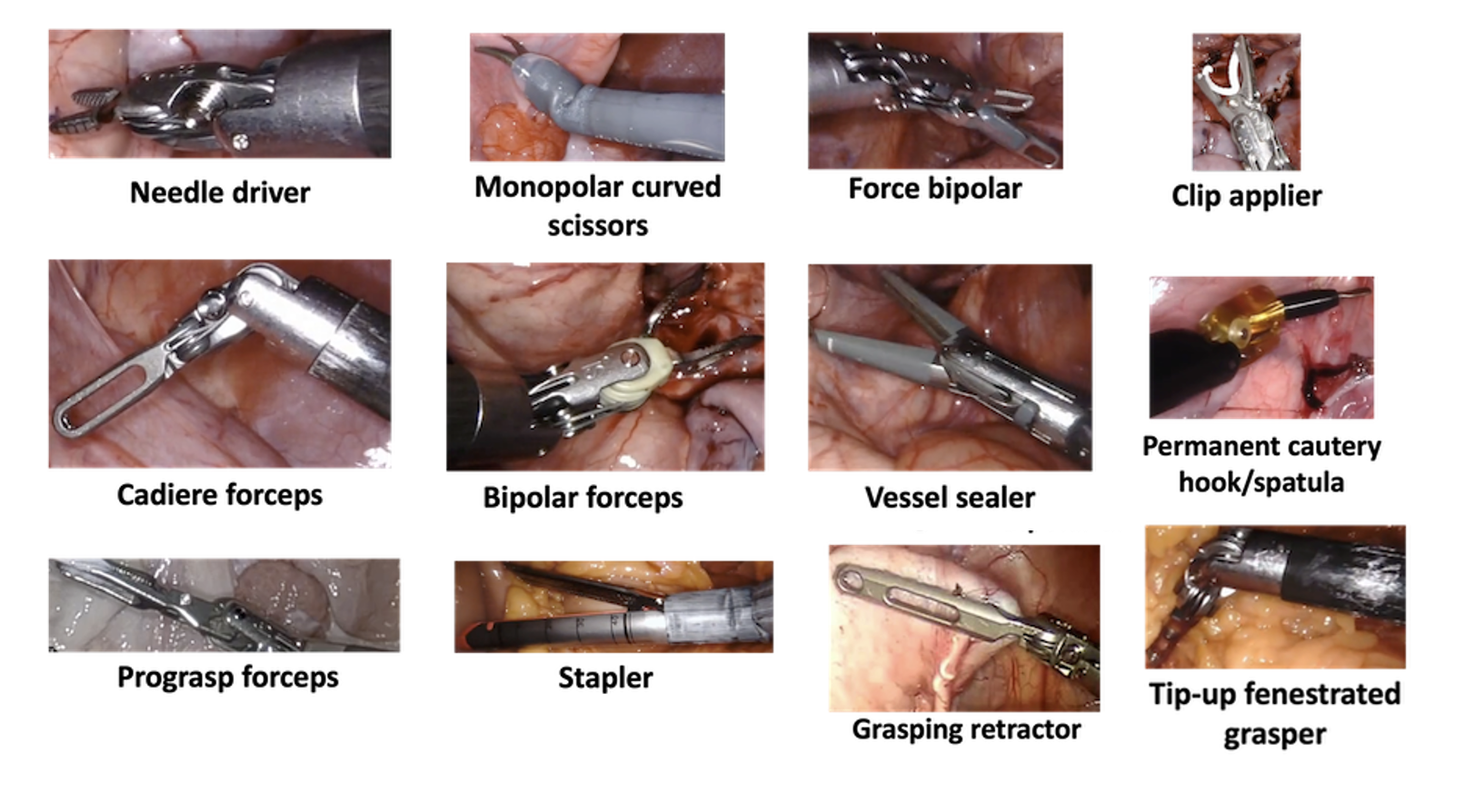}
 \caption{example images of surgical instruments present in the dataset}
  \label{fig:tool_pics}
\end{figure}

Within the recorded surgical clips, up to three robotic surgical tools are present within the surgical field, with up to twelve different tools installed and visible during the surgical exercises. These tools includes: \textit{needle driver, cadiere forceps, prograsp forceps, monopolar curved scissors, bipolar forceps, stapler, force bipolar, vessel sealer, permanent cautery hook/spatula, clip applier, tip-up fenestrated grasper, and grasping retractor} (see Fig \ref{fig:tool_pics}). Additionally, system data indicating the presence of tools was collected in parallel.

\begin{figure}[tbh!]
 \centering
 \includegraphics[scale=0.65]{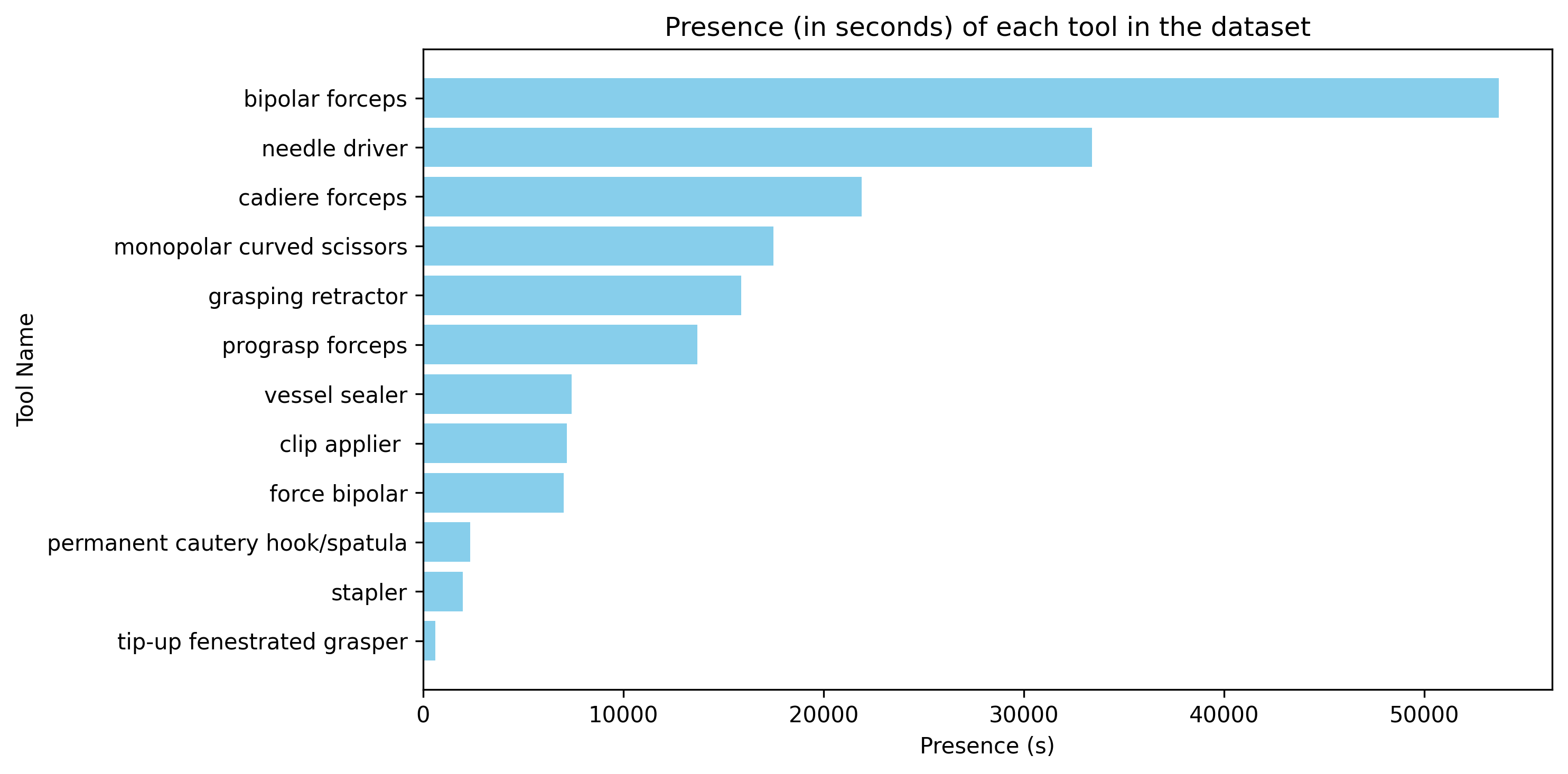}
 \caption{Distribution of tool labels in training dataset}
  \label{fig:total_frames_tools}
\end{figure}

At times surgical tools might be obscured or otherwise temporarily not visible despite being installed. 
Consequently, the tool labels can be considered noisy.
Additionally, owing to the popularity of different instruments, there is a significant imbalance in the distribution of the instruments within the data.
Presence labels can be found in the tools.csv file, with the following format: \textit{install\_case\_part, install\_case\_time, uninstall\_case\_part, uninstall\_case\_time, arm, groundtruth\_toolname.} The distribution of tools within the data is displayed in Fig \ref{fig:total_frames_tools}.\\
\newline
\newline

\noindent{\bf{Steps}}

\begin{figure}
 \centering
 \includegraphics[scale=0.65]{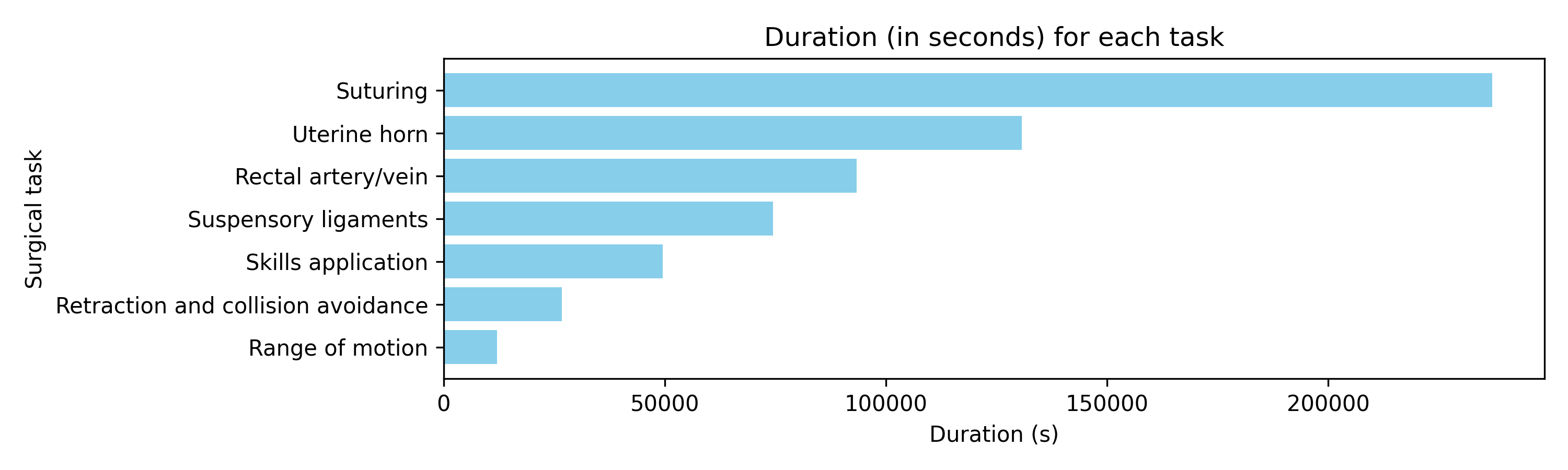}
 \caption{Distribution of surgical steps in training dataset}
  \label{fig:total_seconds_tasks}
\end{figure}

Clinical experts annotated eight surgical training steps executed in a controlled environment. These are: {\it suturing, uterine horn, rectal artery/vein manipulation, suspensory ligaments, general skills application, retraction and collision avoidance, range of motion, and ``other"}.
Step labels can be found in the tasks.csv file, with the following format: \textit{install\_case\_part, install\_case\_time, uninstall\_case\_part, uninstall\_case\_time, arm, groundtruth\_taskname.}
The distribution of surgical steps within the data is displayed below (Fig \ref{fig:total_seconds_tasks}).
\\

\noindent{\bf{Steps Description}}

The dataset labels were expanded to include detailed descriptions of each surgical step, found in the column \texttt{matched\_description} in each \texttt{.csv} file for every case. This addition allows teams to develop vision-language models with more detailed data, specifying what is happening in each step, the anatomical elements involved, the surgeons' behavior, and the tools present.

\begin{table}[h!]
\centering
\caption{Examples of step descriptions.}
\begin{tabular}{p{3cm} p{11cm}}
\hline
\textbf{Step Name} & \textbf{Matched Description} \\
\hline
Suturing & The surgeon will switch the endoscope to 30 degrees down before grasping the inferior portion of the sigmoid colon with the third arm. The surgeon will then retract the arm and lock it into place. The surgeon then marks six rows of two targets each with the monopolar scissors. Then the surgeon will exchange tools with the mega needle driver being placed in the dominant hand and a large needle driver in the non-dominant hand. The surgeon will complete the running suture using both hand-over-hand and pulley methods, with special attention to tissue blanching and lack of movement in the knot. \\[0.5em]
Uterine horn & The surgeon begins by tracing the right ovary and retracting the window of tissue as shown by the trainer. From here, the surgeon will cauterize and transect the fallopian tube, separating it from the uterine horn. The surgeon will then use different energy modalities to mobilize the uterine horn from the broad ligament, continuing medially and downward as far as accessible. These steps will be repeated for the left side. Finally, the surgeon will use both bipolar and monopolar energy to cauterize and amputate the uterine body, freeing it. \\
\hline
\end{tabular}
\end{table}

\noindent{\bf{Questions \& Answers Sample Set}}

The sample set of Q\&A for Visual Language Models (VLMs) contains examples of how the Category 2 for the \href{https://surgvu25.grand-challenge.org/challenge-categories/}{SurgVU25 challenge} was evaluated. For every 30-second video chunk, a sample pair of questions and answers are provided.

For each question, five different answers are available as ground truth. For example, the question:

\begin{quote}
\textit{Was a large needle driver used during the surgery?}
\end{quote}

may have the following list as ground truth answers:

\begin{quote}
\begin{verbatim}
["No", 
 "No, a large needle driver was not used.", 
 "A large needle driver was not used.", 
 "No large needle driver was utilized.", 
 "There is no indication a large needle driver was used."]
\end{verbatim}
\end{quote}
A sample set containing 10 video clips with Q\&A pairs in the evaluation format is available for download here: \url{https://storage.googleapis.com/isi-surgvu/SURGVU25_cat_2_sample_set_public.zip}. 
\\
\\
In summary, the dataset consists of 280 video clips from 155 training sessions, with over 840 hours of surgical steps. All videos were recorded at 60 frames per second, from one channel of the endoscope. This yields approximately 18 million frames at a resolution of 720p (1280 x 720). Annotations on tool presence, surgical step being performed, detailed description of surgical step, and sample question \& answer pairs are also made available.

\subsection*{Validation dataset}

\begin{figure}[tbh!]
 \centering
 \includegraphics[scale=0.45]{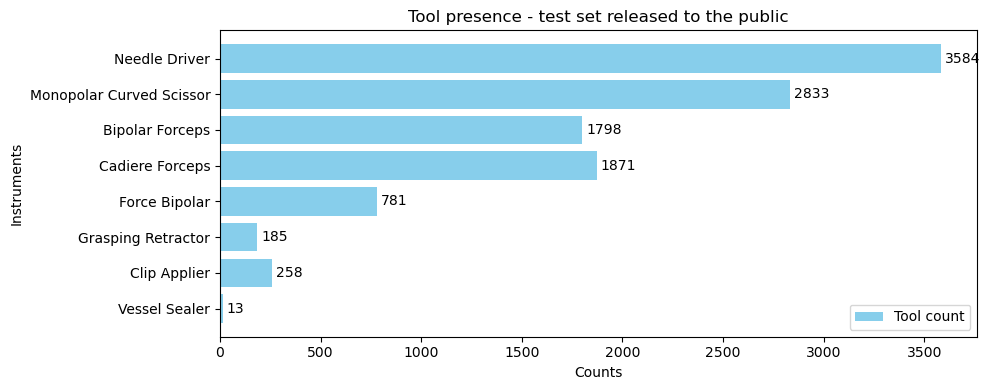}
 \caption{Class distribution of tools in the validation dataset}
  \label{fig:tool_distribution_eval}
\end{figure}

A validation data set (for the tool detection problem from the accompanying MICCAI challenge) was also made available (\url{https://storage.googleapis.com/isi-surgvu/cat1_test_set_public.zip}). It contains video clips with tool-annotated bounding boxes.
The video clips were down sampled to 1 frame per second.
Bounding boxes were annotated by an experienced crowd of annotators. 
Of the twelve tools, 8 are contained in the validation dataset (see Fig \ref{fig:tool_distribution_eval}).

\begin{figure}[tbh!]
 \centering
 \includegraphics[scale=0.5]{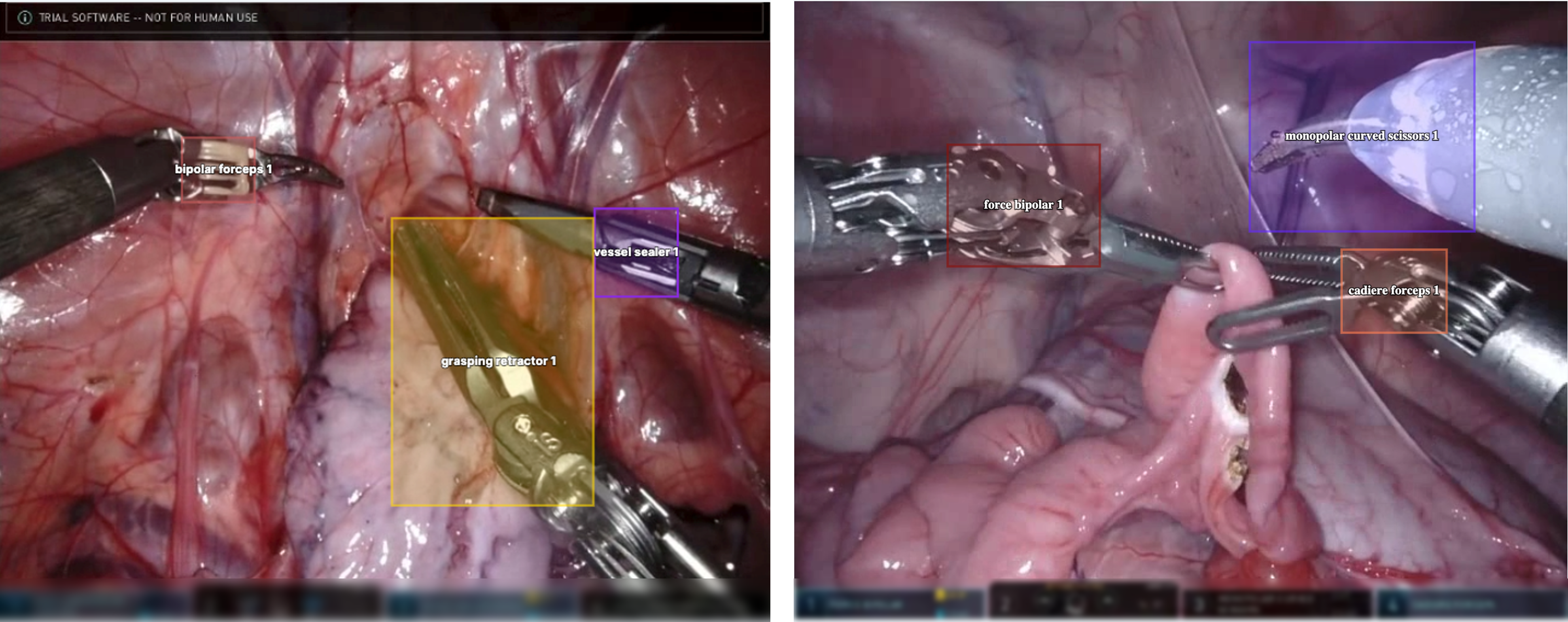}
 \caption{Sample frames with bounding boxes. To prevent the embedded information (description of the tools) from influencing model training and evaluation, the user interface interface was intentionally blurred.}
  \label{fig:test_labels}
\end{figure}

For most tools, the clevis (the rotating wrist joint at the base of the gripping or cutting elements) of the tool is annotated with a bounding box (see examples in Fig \ref{fig:test_labels}). There are two exceptions to this rule:
\begin{itemize}
    \item If the clevis of a tool is not well defined (e.g. the monopolar curved scissors), the tip of the tool is annotated as well.
    \item If the tool is very large and the clevis does not appear in the field of view (e.g. tip-up fenestrated grasper, suction irrigator, stapler, grasping retractor), then the tip of the tool is annotated.
    \newline
    \newline
    \newline
    
\end{itemize}

\section*{Potential Research Topics}

This dataset has the potential to enable many applications in the field of surgical robotics. By no means an exhaustive list, below we have enumerated several active areas.

\begin{itemize}
    \item \textbf{Tracking and segmentation for real-time guidance:} The dataset we are releasing can be employed to develop tracking and segmentation techniques for real-time surgical guidance. In the near future, these techniques can help surgeons to navigate through complex anatomical structures and perform surgical procedures with greater accuracy and precision. For example, by tracking the movement of surgical instruments and segmenting the surgical field, surgeons can get real-time feedback on their actions and make adjustments as needed.
    \item \textbf{Development of weakly supervised/unsupervised methods:} While relatively large, this dataset lacks detailed, noise-free labels. With this data weakly supervised or unsupervised methods can, for instance, enable the autonomous recognition of procedure steps during surgery, or the detection of surgical instruments. Methods that excel under these circumstances would prove invaluable in the many scenarios where obtaining detailed, accurate labels is expensive and laborious. 
    \item \textbf{Activity recognition:} The dataset can be used to develop algorithms that can autonomously recognize procedure steps during surgery. By analyzing the movements of surgical instruments and the surrounding tissue, these algorithms could potentially identify the specific step being performed and provide real-time feedback to the surgeon. This can help reduce the risk of errors and improve surgical outcomes.
    \item \textbf{Video-based performance quantification:} The dataset can be used to quantify surgical performance. By analyzing the movements of surgical instruments and the surrounding tissue, researchers can develop metrics that can objectively evaluate surgical performance. This can help identify areas for improvement and enable surgeons to track their progress over time.
    \item \textbf{Development of video-language models:}
    The dataset contains a variety of surgical activities and anatomy that can be annotated with text captions. This information could be used to train multi-modal embedding models or generative models for applications such as text to video search or video captioning. 
    This can help apply advanced algorithms from the fields of computer vision and natural language to surgical data. 
\end{itemize}

Overall, this dataset has the potential to drive innovation in surgical robotics and improve patient outcomes. By encouraging the development of tracking and segmentation techniques, weakly supervised and unsupervised methods, autonomous recognition of procedure steps, and quantification of surgical performance, the dataset can help improve the accuracy, efficiency, and safety of surgical procedures.

\section*{Discussion}




The SurgVU dataset presents a unique opportunity for advancing surgical data science, with far-reaching implications for clinical applications. Beyond its immediate utility for the accompanying machine learning challenge, this dataset offers a versatile foundation for exploring diverse problems in machine learning, computer vision, and robotics.

One direction for future research involves the employment of pretrained models tailored to the SurgVU dataset. These models can be fine-tuned for various downstream steps, including surgical gesture recognition, tool-tissue interaction analysis, and skill assessment. The availability of pretrained models can significantly reduce the barrier to entry for researchers and facilitate the development of more accurate and efficient algorithms.

Another potential extension of this work involves expanding the dataset through additional annotations or the incorporation of new data sources. Crowdsourcing initiatives or community-driven efforts could provide further labels or annotations, enhancing the dataset's versatility and relevance. For instance, annotating surgical gestures, tool movements, or tissue interactions could enable more sophisticated analysis and modeling of surgical procedures. Furthermore, integrating data from other sources, such as surgical simulations or virtual reality training environments, could broaden the dataset's applicability.

The SurgVU dataset also presents an opportunity for developing video-based objective measures of surgical skill assessment.
By analyzing tool movements, gesture patterns, and procedure execution, researchers may develop accurate and automated evaluation metrics. This could have significant implications for surgical training and quality improvement initiatives.
Future research directions may also explore the application of computer vision and machine learning techniques to specific challenges in surgical data science, such as: tool-tissue interaction analysis for improved safety and precision; surgical gesture recognition for real-time feedback and guidance; procedure recognition and phase detection for enhanced situational awareness; skill assessment and competency evaluation for training and certification.



The SurgVU dataset provides a unique resource for advancing surgical data science. With over 800 hours of video and approximately 18 million labels, it has the potential to become a cornerstone of surgical data science research, driving innovation and collaboration across disciplines. By providing a shared resource for the community, we hope to accelerate progress in robotic-assisted surgery and improve patient outcomes.


\clearpage

\bibliographystyle{unsrt}
\bibliography{references}

\newpage


\end{document}